\documentclass[10pt,twocolumn,letterpaper,table,dvipsnames]{article}

\usepackage{iccv}              

\usepackage{bm}
\usepackage{dsfont}
\usepackage{multirow}
\usepackage[normalem]{ulem}

\usepackage{amsmath}
\DeclareMathOperator*{\argmax}{arg\,max}

\definecolor{iccvblue}{rgb}{0.21,0.49,0.74}
\usepackage[pagebackref,breaklinks,colorlinks,allcolors=iccvblue]{hyperref}

\def\paperID{11153} 
\def\confName{ICCV}
\def\confYear{2025}

\title{Failure Cases Are Better Learned But Boundary Says Sorry: Facilitating Smooth Perception Change for Accuracy-Robustness Trade-Off in Adversarial Training}

\author{Yanyun Wang \hspace{1em} Li Liu\thanks{Corresponding Author: {\tt\small avrillliu@hkust-gz.edu.cn}} \vspace{0.35em} \\
The Hong Kong University of Science and Technology (Guangzhou)
}

\begin{document}
\maketitle
\begin{abstract}
Adversarial Training (AT) is one of the most effective methods to train robust Deep Neural Networks (DNNs). However, AT creates an inherent trade-off between clean accuracy and adversarial robustness, which is commonly attributed to the more complicated decision boundary caused by the insufficient learning of hard adversarial samples. In this work, we reveal a counterintuitive fact for the first time: \textbf{From the perspective of perception consistency, hard adversarial samples that can still attack the robust model after AT are already learned better than those successfully defended.} Thus, different from previous views, we argue that it is rather the over-sufficient learning of hard adversarial samples that degrades the decision boundary and contributes to the trade-off problem. Specifically, the excessive pursuit of perception consistency would force the model to view the perturbations as noise and ignore the information within them, which should have been utilized to induce a smoother perception transition towards the decision boundary to support its establishment to an appropriate location. In response, we define a new AT objective named \textbf{Robust Perception}, encouraging the model perception to change smoothly with input perturbations, based on which we propose a novel \textbf{R}obust \textbf{P}erception \textbf{A}dversarial \textbf{T}raining (\textbf{RPAT}) method, effectively mitigating the current accuracy-robustness trade-off. Experiments on CIFAR-10, CIFAR-100, and Tiny-ImageNet with ResNet-18, PreActResNet-18, and WideResNet-34-10 demonstrate the effectiveness of our method beyond four common baselines and 12 state-of-the-art (SOTA) works. The code is available at \url{https://github.com/FlaAI/RPAT}.
\end{abstract}
\section{Introduction}\label{sec:intro}

\begin{figure}[htbp]
    \centering
    \includegraphics[width=8.3cm]{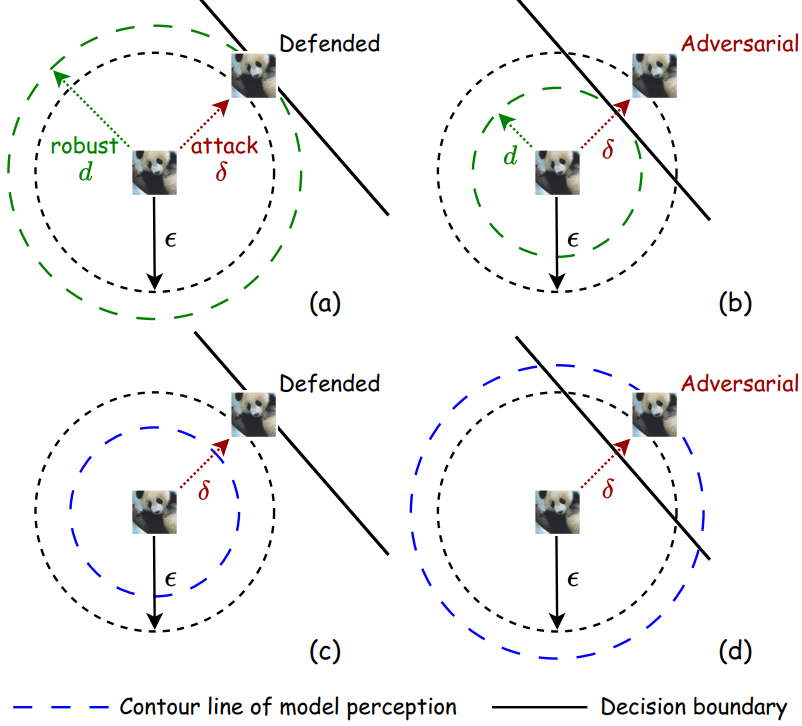}
    \caption{
    Success and failure cases of AT defense from the perspectives of prediction and perception, respectively.
    \textbf{(a)} and \textbf{(b):} AT works through building a robust ball with radius $d$ such that the model prediction remain unchanged within it. Then if the robust $d$-ball fails to contain the $\epsilon$-ball, there would still be failure cases beating the robust model under perturbation $\delta$ within the $\epsilon$-ball. 
    \textbf{(c)} and \textbf{(d):} In this work, we reveal that the smaller $d$-ball in the failure cases is not a result of more vulnerable model perception as previous intuition. In fact, as demonstrated by the contour line, which shows the perturbation ball to the same degree of perception change, the failure cases already achieve less perception change under similar $\delta$ within the $\epsilon$-ball.
    This implies that the issue of these failure cases is not failing to learn perception consistency following the current AT objective, but failing to support the establishment of decision boundary to an appropriate location.
    }
    \label{fig:concept}
\end{figure}

Deep neural networks (DNNs) have demonstrated impressive performance across various real-world applications~\cite{liang2024merge,zhou2024sdformer,lin2025backdoordm}. However, they are inherently vulnerable to adversarial attacks~\cite{biggio2013evasion, szegedy2014intriguing}. Specifically, even imperceptible perturbations to inputs can cause substantial changes in the outputs~\cite{goodfellow2015explaining, huang2020survey}. As this problem raises significant security concerns~\cite{wang2020improving}, previously, several defense strategies, such as Defense Distillation~\cite{papernot2017practical}, Feature Squeezing~\cite{xu2017feature}, Randomization~\cite{xie2018mitigating}, and Input Denoising~\cite{guo2018countering, liao2018defense}, have been proposed to improve the adversarial robustness of DNNs. However, most of these approaches have proven ineffective against more advanced adaptive attacks~\cite{athalye2018obfuscated, tramer2020adaptive}, such as Auto-Attack~\cite{croce2020reliable}. Currently, Adversarial Training (AT)~\cite{goodfellow2015explaining, madry2018towards} is regarded as one of the most effective methods for training robust DNNs~\cite{athalye2018obfuscated, dong2020benchmarking}. 

However, an \textbf{accuracy-robustness trade-off} problem have been widely identified for AT in recent studies  ~\cite{tsipras2019robustness,zhang2019theoretically,raghunathan2019adversarial,raghunathan2020understanding,wang2020once,bai2021recent,rade2022reducing,yin2023push,wang2024new,liu2025parameter}. Compared with standard training, the improvement in adversarial robustness acquired by AT is usually at the expense of the reduction in clean accuracy, which is afraid to significantly degrade the user experience in clean context and hinder the application of AT in real-world practices. The current consensus is that AT aggravates the accuracy-robustness trade-off by creating overly complex decision boundaries, primarily because models struggle to sufficiently learn from hard adversarial samples during training~\cite{ding2020mma,liu2021probabilistic,dong2022exploring,rade2022reducing,cheng2022cat,yang2022one,kanai2023one,yin2023push,wang2024new,liu2025parameter}. Specifically, as the conventional AT objective encourages the model to perceive any adversarial inputs within the $\epsilon$-ball similar to the corresponding benign one to produce consistent predictions for them, previous works empirically assume that the wrong predictions of the hard adversarial samples are attributed to the insufficiently optimized perception of them under the current objective. Therefore, although defining ``hard sample'' from different perspectives (\eg, classification result and distance to decision boundary), previous works agree to further optimize the perception consistency on them, ending up with various learning strategies specific for them, such as data reweighting~\cite{zhang2021geometry,liu2021probabilistic}, adaptive perturbation radius~\cite{ding2020mma,rade2022reducing}, and non-one-hot supervised signal~\cite{dong2022exploring,kanai2023one}.

However, in this work, we reveal that for the robust model trained by convention AT, \textbf{the statistical difference in model perception from benign to hard adversarial samples causing failure defense is already smaller than to others (\ie, those successfully defended)}. It implies that the model has even more sufficiently learned them under the current AT objective, and should have successfully defended them if the above assumption holds. Thus, as demonstrated in Figure~\ref{fig:concept}, we suggest that the key factor here is no longer the optimization of model perception but the location of the decision boundary. In that case, keeping on pursuing over-sufficient small perception change would not only help little in further reducing the prediction error, but also force the model to view the perturbations as noise and ignore the information within them, which impairs their ability in claiming robust balls with reasonable gradients to push the decision boundary to an appropriate location. As a consequence, such sample-level imbalance in boundary establishment would naturally increase its complexity, harming generalization ability and degrading the trade-off between accuracy and robustness.

To mitigate the over-sufficient learning of model perception on the failure cases and retain effective gradient signals \textit{w.r.t.} them to facilitate the establishment of the decision boundary to an appropriate location, \textbf{firstly}, we define a new AT objective named \textbf{Robust Perception} in addition to the conventional one, encouraging smoother perception changes towards the decision boundary along with the input perturbations. Rigorous theoretical derivations also demonstrate its effectiveness in smoothening the decision boundary and releasing the current trade-off problem from the perspectives of local linearity~\cite{goodfellow2015explaining} and \textit{Lipschitz} regularization~\cite{krishnan2020lipschitz,pauli2021training,fazlyab2023certified,abuduweili2024estimating}. \textbf{Secondly}, based on our new objective, we propose a novel \textbf{R}obust \textbf{P}erception \textbf{A}dversarial \textbf{T}raining (\textbf{RPAT}) method, concurrently improving clean accuracy and adversarial robustness in AT. Extensive experiments on CIFAR-10, CIFAR-100, and Tiny-ImageNet with ResNet-18, PreActResNet-18, and WideResNet-34-10 demonstrate the advancement of RPAT beyond four common benchmarks and 12 state-of-the-art (SOTA) works. 

\section{Background and Related Works}\label{sec:backg}

\subsection{Robustness and Adversarial Training}\label{subsec:backg_AT}

Adversarial robustness reflects model vulnerability to malicious perturbations, which is commonly quantified by test accuracy under adversarial attacks~\cite{szegedy2014intriguing,bai2021recent,wang2022tsfool}. Representatively, the Fast Gradient Sign Method (FGSM)~\cite{goodfellow2015explaining} perturbs input samples along the gradient of the loss function, whereas the Projected Gradient Descent (PGD) attack~\cite{madry2018towards} applies FGSM iteratively and projects the perturbation onto the $\epsilon$-ball. Auto-Attack~\cite{croce2020reliable} introduces and combines several parameter-free attacks to mitigate practical issues such as hyper-parameter tuning and gradient masking. Due to their effectiveness, PGD and Auto-Attack are widely used to assess adversarial robustness, and so does this work. 

Previously, although many defense strategies including Defense Distillation~\cite{papernot2017practical}, Feature Squeezing~\cite{xu2017feature}, Input Denoising~\cite{guo2018countering,liao2018defense} and Randomization~\cite{xie2018mitigating} have been developed, most rely on obfuscated gradients~\cite{athalye2018obfuscated} and can be evaded by advanced adaptive attacks~\cite{tramer2020adaptive}. Nowadays, Adversarial Training~\cite{goodfellow2015explaining,madry2018towards} is recognized as the most effective method to enhance adversarial robustness~\cite{athalye2018obfuscated,dong2020benchmarking,tao2021better}. Unlike clean training on benign samples, AT directly learns on adversarially augmented data. Its objective can be formulated as the following $\min$-$\max$ problem~\cite{madry2018towards}: \begin{equation}\label{eq:AT}
    \min_{\bm{\theta}} \frac{1}{n} \hspace{0.1em} \sum_{i=1}^n \max_{\hspace{0.1em} \Vert \mathbf{x}_i' - \mathbf{x}_i \Vert_{p} \leq \epsilon \hspace{0.1em}} \mathcal{L}(f_{\bm{\theta}}(\mathbf{x}_i'), y_i),
\end{equation}
where $f_{\bm{\theta}}(\mathbf{x}_i) = \argmax_{c = 1, ..., C} \hspace{0.1em} \mathbf{p}_{c} (\mathbf{x}_i, \bm{\theta})$ with $\mathbf{p}_{c}$ denoting the \textit{softmax} probability of class $c$ within the prediction of $\mathbf{x}_i$, $n$ denotes the number of training samples, and $\mathbf{x}_i'$ is generated by adding the strongest $L_p$-norm perturbation within the $\epsilon$-ball to the original sample $\mathbf{x}_i$. This objective aims at prediction consistency within the $\epsilon$-ball, encouraging the model to also produce similar perception there.

PGD-AT~\cite{madry2018towards} is the first attempt at solving this $\min$-$\max$ optimization iteratively by approximating the inner maximization via PGD to craft adversarial samples and then performing stochastic gradient descent (SGD) on these samples for the outer minimization. TRADES~\cite{zhang2019theoretically} first considers balancing clean accuracy and adversarial robustness in AT, which decomposes robust error into natural error and boundary error, with the latter occurring only if specific data points are sufficiently close to the decision boundary. MART~\cite{wang2020improving} further refines AT by distinguishing misclassified from correctly classified samples, which suggests appending boundary error only for misclassified samples to the basic AT objective in PGD-AT. Consistency-AT~\cite{tack2022consistency} encourages the prediction between adversarial samples derived from different augmentations of the same benign sample to be similar, which explicitly enhances the prediction consistency further.

\subsection{Accuracy-Robustness Trade-Off in AT}\label{subsec:backg_tradeoff}

An important problem identified for AT is the trade-off between clean accuracy and adversarial robustness~\cite{tsipras2019robustness,zhang2019theoretically,raghunathan2019adversarial,raghunathan2020understanding,wang2020once,bai2021recent,rade2022reducing,yin2023push,liu2025parameter}, which means the robustness gained by AT often comes at the cost of reduced model accuracy compared with clean training. Consequently, AT can degrade user experience under benign conditions, which may seriously delay its real-world deployment. A widely acknowledged cause of this problem is that AT tends to learn more complicated decision boundaries than clean training~\cite{dong2022exploring,rade2022reducing,cheng2022cat,yang2022one,yin2023push,liu2025parameter}, thereby harming the ability of the robust model to generalize to unseen data. A number of previous works attribute this to the insufficient learning of hard adversarial samples~\cite{ding2020mma,zhang2021geometry,liu2021probabilistic,rade2022reducing,dong2022exploring,kanai2023one}. Though defining ``hard adversarial samples'' from different perspectives, such as the classification result and the distance to decision boundary, they basically reached a consensus that the model perception \textit{w.r.t.} hard adversarial samples is not well-optimized, and as a consequence, fails to produce the consistent final prediction. In response, these works propose various solutions to help the model learn better prediction consistency for those hard adversarial samples.

For instance, GAIRAT~\cite{zhang2021geometry} proposes that hard samples nearer to the decision boundary are more critical but less robust and should be assigned a larger weight. MAIL~\cite{liu2021probabilistic} further proposes three types of probabilistic margins to quantify the closeness for better reweighting adversarial data accordingly. In contrast, MMA~\cite{ding2020mma} suggests learning hard adversarial samples with reduced perturbations, which employs an adaptive $\epsilon$ for perturbations to maximize the margin between data and the decision boundary. EWAT~\cite{kim2021entropy} weighs the loss for each adversarial sample proportionally to the entropy of its prediction distribution during AT to focus on those with more uncertain labels. TE~\cite{dong2022exploring} reveals that the model may attempt to memorize certain hard adversarial samples during AT, as it is difficult to assign high-confidence one-hot labels for them~\cite{stutz2020confidence,cheng2022cat}. SOVR~\cite{kanai2023one} suggests increasing the margins of \textit{logits} for certain hard samples by switching to a one-vs-the-rest loss.

Additionally, several works also study the trade-off problem from other perspectives. AWP~\cite{wu2020adversarial} proposes a double perturbation mechanism that flattens the loss landscape by weight perturbation to improve robust generalization. KD+SWA~\cite{chen2021robust} injects more smoothening during AT by leveraging knowledge distillation to smooth the logits and performing stochastic weight averaging~\cite{izmailov2018averaging} to smooth the weights. ADR~\cite{wu2024annealing} employs soft labels as a guidance mechanism, accurately reflecting the distribution shift under attack during AT. CURE~\cite{gowda2024conserve} selectively updates specific layers while preserving others to enhance the learning capacity of the model. ReBAT~\cite{wang2024balance} views AT as a dynamic min-max game between the model trainer and the attacker, rebalancing the two players by either constraining the trainer’s capacity or intensifying the attack.

\section{Motivation: AT Over-Learns Failure Cases}\label{sec:motivation}

It is basically common sense for DNN-based classification that the model learns to perceive input better and is thus able to predict it more precisely (especially when the model perception is indicated by its \textit{logits}). Therefore, it seems reasonable at first glance that the previous works assume that the wrong prediction of the hard adversarial samples is due to the insufficient model perception of them, and propose various solutions upon this assumption in response. 

However, to the best of our knowledge, there is no previous discussion on a simple but crucial question: \textbf{What if the assumption does not hold?} If the model perception on the hard adversarial samples is already relatively close to the corresponding benign ones (\eg, at a similar or even better level than non-hard ones), then it would be no longer effective enough to further optimize the model on such samples through an objective simply pursuing prediction consistency as the conventional AT. Also, in turn, this can no longer effectively benefit the further reduction of the prediction error. Surprisingly, our following proof-of-concept experiments reveal that this is exactly the case.

\begin{figure}[htbp] 
    \centering
    \includegraphics[width=7.64cm]{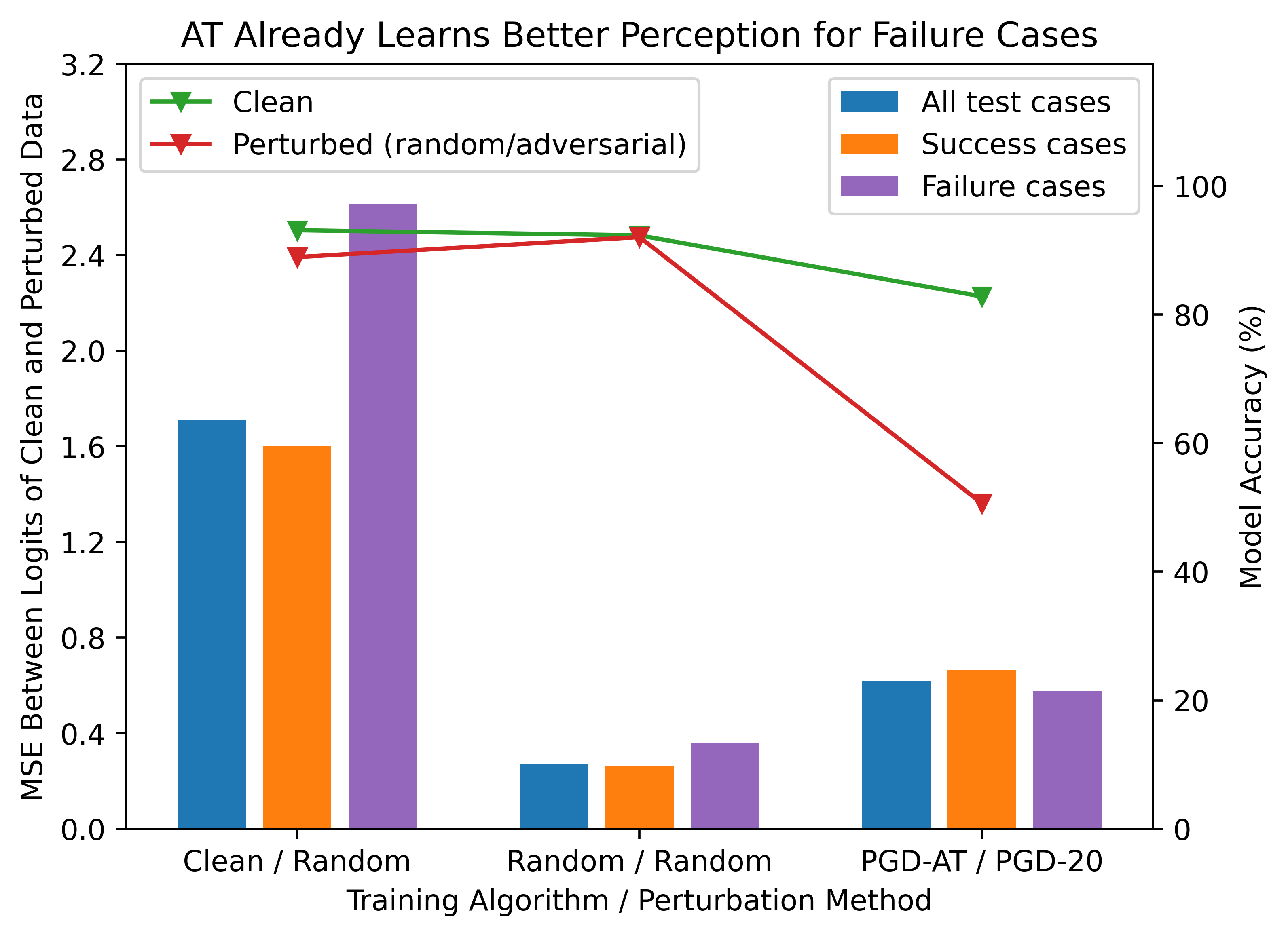}
    \caption{Proof-of-concept experiments for the different learning sufficiency in model perception for success and failure cases on CIFAR-10 with ResNet-18. In the x-axis, the ``Random'' for ``Training Algorithm'' means the training data is randomly perturbed within the budget $\epsilon$, while that for ``Perturbation Method'' indicates the attack used for evaluation on test data is random perturbation. The bars demonstrate the \textit{MSE} score on \textit{logits} between each pair of benign and perturbed samples, with the lines additionally showing the model accuracy on clean and perturbed data. Our results reveal a surprising phenomenon that, different from the clean training (with or without random noise), where failure cases are worse learned as we intuitively expected, AT already learns a better model perception for failure cases.}
    \label{fig:motivation}
\end{figure}

The proof-of-concept experiments are conducted on CIFAR-10 dataset with ResNet-18 architecture, with the results illustrated in Figure~\ref{fig:motivation}. As suggested by the x-axis, we involve three cases with different pairs of training and evaluation methods, which are respectively 1) trained on clean data and evaluated with random perturbations, 2) trained and evaluated with randomly perturbed data, and 3) trained with PGD-AT and evaluated with PGD-20 adversary. The budget of the random perturbations $\epsilon \!=\! 8/255$ is assigned the same as PGD, following our experimental setup detailed in Appendix~\ref{subsec:exp_setup}, so do other training and evaluation settings. Given that the conventional AT objective in Equation~(\ref{eq:AT}) encourages the model to produce a similar perception for benign and adversarial samples, we utilize the similarity in model perception between the corresponding benign and perturbed samples to estimate the learning sufficiency of the perturbed samples in AT. Specifically, the model perception is represented by the \textit{logits}, and the similarity is measured through the \textit{mean squared error (MSE)} score. We calculate the score respectively for success and failure cases of defense, which are divided according to the correctness of the final classification results of the perturbed samples. In addition, the model accuracy on clean and perturbed data is also recorded and illustrated together.

Surprisingly, we experimentally found that \textbf{the failure cases already learn better in model perception than the success cases}. First, for the clean trained model, the perception on failure cases is significantly worse than on the success ones, just as we expected. Although the difference between success and failure cases has decreased for the randomly trained model, the failure cases remain worse, which suggests that non-adversarial perturbations are influential but not enough to change the trend. However, unlike the previous intuition, the model trained by PGD-AT can already perceive failure cases better than success ones.
Our finding updates the current knowledge that, unlike those in clean training, the failure cases in AT do not fail simply because they are not sufficiently learned. On the contrary, the model even learns them harder to try to achieve the prediction and perception consistency. The smaller perception change under similar perturbations is direct evidence for this point of view. 

Then, as a straightforward deduction, now that we have success cases with worse perception but failure cases with better perception, there should be a statistical difference between them in the maximum perception change they can tolerate while still maintaining the consistent prediction, just as demonstrated in Figure~\ref{fig:concept}. In other words, the decision boundary is established relatively closer to the failure cases, which is instead the issue we should focus on, because such sample-level imbalance in boundary establishment can naturally complicate it and thus degrade the accuracy-robustness trade-off as mentioned in Section~\ref{subsec:backg_tradeoff}. In that case, the over-consistent model perception between benign and adversarial samples in the failure cases rather harms the final robustness, because this is expected to force the model to view the perturbations as noise and ignore the information within them, which should have been utilized to induce a smoother perception transition towards the decision boundary to support its establishment, thus impairing the ability of such cases to claim a reasonable robust distance during AT compared with success cases. Therefore, in this work, we argue that \textbf{it is not insufficient but over-sufficient learning of failure cases that contributes to the more complicated decision boundary and finally results in the current trade-off problem}.


\section{Proposed Method}\label{sec:method}


\begin{figure*}[htbp]
    \centering
    \includegraphics[width=17cm]{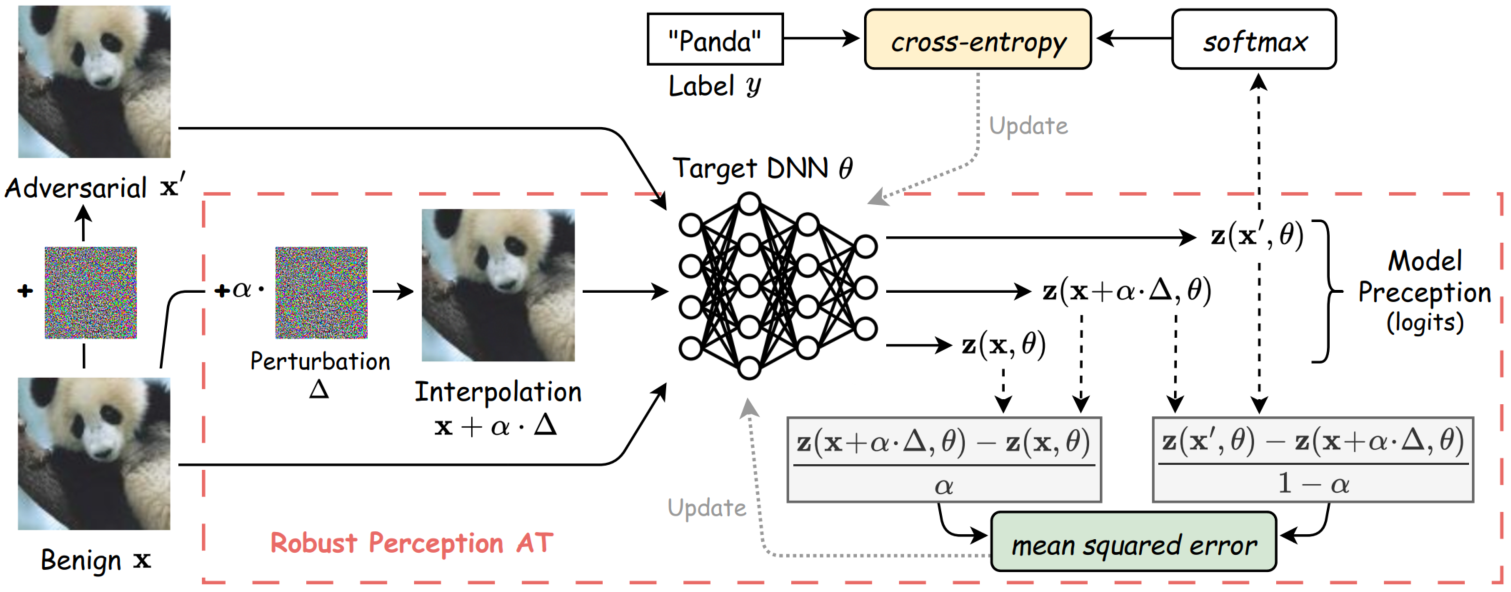}
    \caption{Illustration of the proposed RPAT method. In addition to the conventional AT objective, which learns adversarial samples with the original supervised labels, RPAT introduces a novel regularization term to achieve the newly defined \textbf{\textit{Robust Perception}} objective. 
    Given benign sample $\mathbf{x}$ and adversarial samples $\mathbf{x}' = \mathbf{x} + \Delta$, for any interpolation samples $\mathbf{x} + \lambda \cdot \Delta$ with $\lambda \in [0, 1]$, it encourages the ratio of the perception changes, $z_{\bm{\theta}}(\mathbf{x}+\lambda\cdot\Delta)-z_{\bm{\theta}}(\mathbf{x})$ and $z_{\bm{\theta}}(\mathbf{x}')-z_{\bm{\theta}}(\mathbf{x}+\lambda\cdot\Delta)$, respectively from $\mathbf{x}$ to $\mathbf{x} + \lambda \cdot \Delta$ and from $\mathbf{x} + \lambda \cdot \Delta$ to $\mathbf{x}'$, to approach the ratio indicated by $\lambda$, such that the model perception would change smoothly along with the input perturbations from $\mathbf{x}$ to $\mathbf{x}'$.}
    \label{fig:roadmap}
\end{figure*}

\subsection{Robust Perception: A New AT Objective}\label{subsec:objective}

In Section~\ref{sec:motivation}, the too-small perception change under perturbations caused by the over-sufficient learning of failure cases has been identified as an important factor contributing to the current trade-off problem, which in turn questions the effectiveness of the conventional AT objective on such failure cases. 

Formally, under Equation~(\ref{eq:AT}) which implicitly optimize the perception distance $\| h_{\bm{\theta}}(\mathbf{x}') - h_{\bm{\theta}}(\mathbf{x}) \| \to 0$ between corresponding benign sample $\mathbf{x}$ and adversarial sample $\mathbf{x}'$, the gradient of the conventional AT loss \textit{w.r.t.} the perturbation, $\nabla_{\| \mathbf{x}' - \mathbf{x} \|}\mathcal{L}$, would also approach $0$. In that case, the model cannot further optimize the decision boundary using the perturbation information. Thus, a straightforward idea to mitigate this is to encourage smoother perception change towards the decision boundary along with the input perturbation, such that $\| h_{\bm{\theta}}(\mathbf{x}') - h_{\bm{\theta}}(\mathbf{x}) \| \propto \| \mathbf{x}' - \mathbf{x} \|$, as long as the final prediction remains unchanged. This is expected to retain an effective gradient signal in the direction of the perturbation, which enables the model to utilize this information to guide the decision boundary to a more reasonable location. Accordingly, in this section, we propose \textit{Robust Perception} as a new AT objective in addition to the conventional one to realize this idea.

\vspace{0.5em} \noindent \textbf{Definition 1} (\textit{\textbf{Robust Perception}})\textbf{.} \hspace{1pt} \textit{Provided an AT task with target model $\bm{\theta}$, let $\mathbf{x}$ and $\mathbf{x}'$ be any pairs of corresponding benign and adversarial samples with $\Delta = \mathbf{x}' - \mathbf{x}$, for any hidden representation $h_{\bm{\theta}}(\mathbf{x})$ denoting the model perception, our additional AT objective can be formulated as:}
\begin{equation*}
\forall \, \alpha \in [0, 1], \, \| h_{\bm{\theta}}(\mathbf{x} + \alpha \cdot \Delta) - h_{\bm{\theta}}(\mathbf{x}) \| = \alpha \cdot \| h_{\bm{\theta}}(\mathbf{x}') - h_{\bm{\theta}}(\mathbf{x}) \|.
\end{equation*}

\vspace{0.1em} To facilitate the understanding of the proposed \textit{Robust Perception}, in addition to the most straightforward perspective that it encourages the change in model perception along with the adversarial perturbations, the effectiveness of this new objective can also be demonstrated from other two perspectives, local linearity~\cite{goodfellow2015explaining} and \textit{Lipschitz} regularization~\cite{krishnan2020lipschitz,pauli2021training,fazlyab2023certified,abuduweili2024estimating}, both of which also help smooth the decision boundary and mitigate the current trade-off problem. Below, we further detail them theoretically.

\vspace{0.5em} \noindent \textbf{Theorem 1.} \textit{Let H be the Hessian Matrix such that $H_{h_{\bm{\theta}}}(\mathbf{x})$ $= \nabla_\mathbf{x}^2h_{\bm{\theta}}(\mathbf{x})$, then with the new optimization objective of Robust Perception, we have:}
\begin{equation*}
\forall \, \Delta, \,\, \Delta^\top \!\cdot H_{h_{\bm{\theta}}}(\mathbf{x}) \cdot \Delta \to 0.
\end{equation*}

\noindent \textit{Proof.} We defer the proof to Appendix~\ref{app:theorem_1}.

\vspace{0.5em} This means \textit{Robust Perception} limits the second-order (and higher-order as well) nonlinear effects within the adversarial perturbation to the model perception, such that the perception would change mainly along with the linear term of the perturbation, and thus be smoother. This also aligns with the local linearity hypothesis of adversarial samples~\cite{goodfellow2015explaining}, which assumes that linear approximations dominate the model behavior under adversarial perturbations. 

\vspace{0.5em} \noindent \textbf{Theorem 2.} \textit{Let J be the Jacobian Matrix such that $J_{h_{\bm{\theta}}}(\mathbf{x})$ $= \nabla_\mathbf{x} h_{\bm{\theta}}(\mathbf{x})$, then with the new optimization objective of Robust Perception, we have:}
\begin{equation*}
\forall \, \alpha \in [0,1], \,\, J_{h_{\bm{\theta}}}(\mathbf{x} + \alpha \cdot \Delta) \to J_{h_{\bm{\theta}}}(\mathbf{x}),
\end{equation*}
\textit{then with $\| \cdot \|_{\text{spec}}$ denoting the Spectral Norm and $\gamma$ being any micro value, given $\| J_{h_{\bm{\theta}}}(\mathbf{x} + \alpha \cdot \Delta) - J_{h_{\bm{\theta}}}(\mathbf{x}) \|_{\mathrm{spec}} \leq \gamma$, the function $h_{\bm{\theta}}(\mathbf{x})$ can be referred to as $K$-Lipschitz with the Lipschitz constant $K$ upper-bounded by:}
\begin{equation*}
K \,\leq\, \sup_\mathbf{x} \| J_{h_{\bm{\theta}}}(\mathbf{x}) \|_{\mathrm{spec}} + \gamma.
\end{equation*}

\noindent \textit{Proof.} We defer the proof to Appendix~\ref{app:theorem_2}.

\vspace{0.5em} This theorem tells us that \textit{Robust Perception} also regularizes the change $\gamma$ in \textit{Jacobian} along with the adversarial perturbation, which can be interpreted as limiting the increase of the global \textit{Lipschitz} constant by $\gamma$ under perturbation. Previous works have demonstrated that this is expected to help achieve smoother decision boundaries and improve the accuracy-robustness trade-off~\cite{krishnan2020lipschitz,pauli2021training,fazlyab2023certified,abuduweili2024estimating}.

\subsection{Robust Perception Adversarial Training}\label{subsec:RPAT}

Based on the above considerations, in this section, we propose a novel AT method named \textbf{R}obust \textbf{P}erception \textbf{A}dversarial \textbf{T}raining (\textbf{RPAT}). It utilizes a new additional regularization term appended to the conventional AT objective to achieve the \textit{Robust Perception}. Formally, given a $C$-class classification task ($C \geq 2$) and a dataset $\mathcal{D} = \{(\mathbf{x}_i, y_i)\}_{i = 1, ..., n}$ with benign sample $\mathbf{x}_i \in \mathbb{R}^{d}$ and supervised label $y_i \in \{1, ..., C\}$, let $\mathbf{\hat{x}}_i'$ denotes the adversarial sample corresponding to $\mathbf{x}_{i}$, which is generated as $\mathbf{\hat{x}}_i' = \argmax_{\Vert \mathbf{x}_i' - \mathbf{x}_i \Vert_{p} \leq \epsilon} \mathds{1}(\argmax_{k = 1, ..., C} h_{\bm{\theta}}(\mathbf{x}_i')^{(k)} \neq y_i)$, then based on the Definition 1 in Section~\ref{subsec:objective}, the adversarial risk to be minimized for the proposed RPAT \textit{w.r.t.} $\mathbf{x}_{i}$ can be formulated with the \textit{0-1 loss}~\cite{zhang2019theoretically} as:
\begin{equation}\label{eq:risk}
\begin{aligned}
    & \mathcal{R}(\bm{\theta}, \mathbf{x}_i) := \, \mathds{1}(f_{\bm{\theta}}(\mathbf{\hat{x}}_i') \neq y_i) \\ 
    & \hspace{2.4em} + \, \mathds{1} \Big( \frac{h_{\bm{\theta}}(\tilde{\mathbf{x}}_i) - h_{\bm{\theta}}(\mathbf{x}_i)}{\alpha} \neq \frac{h_{\bm{\theta}}(\mathbf{\hat{x}}_i') - h_{\bm{\theta}}(\tilde{\mathbf{x}}_i)}{1 - \alpha} \Big),
\end{aligned}
\end{equation}
where $\tilde{\mathbf{x}}_i = \mathbf{x}_i + \alpha \cdot (\mathbf{\hat{x}}_i' - \mathbf{x}_i)$.
 
While the proposed \textit{Robust Perception} aims to improve learning efficacy for the failure cases, the above risk does not explicitly distinguish success and failure cases.
This is because \textit{Robust Perception} is designed to mitigate the gap of the failure cases in the effectiveness of learning compared with the success ones, rather than require any new characteristic beyond the success cases. In other words, the success cases are expected to also satisfy \textit{Robust Perception} at the beginning. Therefore, considering the computational efficiency, it is not necessary to specifically exclude the success cases for the \textit{Robust Perception} term.

Finally, by introducing practical surrogate losses to optimize the adversarial risk in Equation~(\ref{eq:risk}), we end up with our new RPAT method. To be specific, for the supervised learning term for adversarial samples, we adopt \textit{cross-entropy (CE)} just as the conventional AT by default. Then, for the \textit{Robust Perception} term, as it is formulated in a format of similarity regularization, various criterion that measures the similarity distance between the two input items are feasible options, such as \textit{mean squared error (MSE)}, \textit{Kullback–Leibler divergence (KL)}, \textit{Jensen-Shannon divergence (JS)}, and \textit{Cosine similarity}. We simply adopt \textit{MSE loss} for the basic version of RPAT here, and would also demonstrate the effectiveness of \textit{KL loss} later in Section~\ref{subsec:exp_SOTAs}. With the \textit{logits} $\mathbf{z}(\mathbf{x}_i,\bm{\theta})$ utilized as the hidden representation $h_{\bm{\theta}}(\cdot)$ to denote the model perception, and $\mathbf{p} (\mathbf{x}_i, \bm{\theta})$ being the \textit{softmax} output as given in Equation~(\ref{eq:AT}), the \textbf{overall objective} of RPAT can be formulated as:
\begin{equation}\label{eq:RPAT}
{\footnotesize
\begin{aligned}
    & \mathcal{L}^{\textit{RPAT}}(\bm{\theta}, \mathcal{D}, \lambda, \alpha) := \frac{1}{n} \sum_{i=1}^{n} \bigg( \mathcal{L}^{\textit{CE}}(\mathbf{p}(\hat{\mathbf{x}}_i', \bm{\theta}), y_i) \\ 
    & \hspace{1.69em} 
    +\lambda\cdot\mathcal{L}^{\textit{MSE}}\Big( \frac{\mathbf{z}(\tilde{\mathbf{x}}_i,\bm{\theta}) - \mathbf{z}(\mathbf{x}_i,\bm{\theta})}{\alpha} \Big|\Big| \frac{\mathbf{z}(\hat{\mathbf{x}}_i',\bm{\theta}) - \mathbf{z}(\tilde{\mathbf{x}}_i,\bm{\theta})}{1 - \alpha} \Big) \bigg),
\end{aligned}
}
\end{equation}
where $\lambda$ is a weight hyper-parameter to balance the conventional AT objective and our new \textit{Robust Perception} objective. An intuitive illustration of the proposed RPAT method is provided in Figure~\ref{fig:roadmap}.

\section{Experiments}\label{sec:exp}

Following previous AT works, the effectiveness of the proposed RPAT is evaluated on CIFAR-10, CIFAR-100~\cite{krizhevsky2009learning} and Tiny-ImageNet~\cite{li2015tiny} datasets. The main experiments can be divided into two parts. First, based on Tack et al.~\cite{tack2022consistency}, we implement RPAT upon three common AT baselines, namely PGD-AT, TRADES, and MART, as well an advanced method, Consistency-AT, with ResNet-18~\cite{he2016deep} architecture. This is to show the general effectiveness of the proposed \textit{Robust Perception} in improving the current trade-off. Second, by integrating RPAT into one of the current SOTA methods, ReBAT~\cite{wang2024balance}, with PreActResNet-18~\cite{he2016identity} and WideResNet-34-10~\cite{zagoruyko2016wide} architectures, we achieve new SOTA performance beyond 12 SOTA methods on the accuracy-robustness trade-off problem. This is to demonstrate the advancement achieved by the proposed RPAT in the AT area. The detailed experimental setup is provided in Appendix~\ref{subsec:exp_setup}.

\begin{table*}[htb]
  \caption{Improvements that the proposed RPAT achieved upon four benchmarks. The results are acquired with ResNet-18 and $\ell_{\infty}$ norm.}
  \label{tab:l_inf}
  \centering
  \renewcommand\arraystretch{0.95}
  \resizebox{17.47cm}{!}{
    \begin{tabular}{llccccccccccccc}
    \toprule
    \multicolumn{1}{c}{\multirow{3.6}{*}{Dataset}} & \multicolumn{1}{c}{\multirow{3.6}{*}{Method}} & \multicolumn{5}{c}{\textbf{Benchmarks}} & \multicolumn{8}{c}{\textbf{+RPAT (Ours)}} \\ 
    \cmidrule(r){3-7} \cmidrule(r){8-15}
    \multicolumn{1}{c}{} & \multicolumn{1}{c}{} & Clean & PGD-20 & AA & \multirow{2.3}{*}{Mean} & \multirow{2.3}{*}{NRR} & \multicolumn{2}{c}{Clean} & \multicolumn{2}{c}{PGD-20} & \multicolumn{2}{c}{AA} & \multirow{2.3}{*}{Mean} & \multirow{2.3}{*}{NRR} \\ 
    \cmidrule{3-3} \cmidrule{4-4} \cmidrule{5-5} \cmidrule{8-9} \cmidrule{10-11} \cmidrule{12-13}
    \multicolumn{1}{c}{} & \multicolumn{1}{c}{} & best & best & best &  &  & best & final & best & final & best & final &  &  \\ 
    \midrule
    \multirow{4}{*}{CIFAR-10} & PGD-AT & 82.92 & 50.61 & 46.74 & 64.830 & 59.782 & 83.20 & 82.99 & 51.29 & 50.97 & 48.00 & 47.60 & 65.600 & 60.878 \\
     & TRADES & 79.67 & 51.91 & \underline{47.62} & 63.645 & \underline{59.610} & 80.02 & 80.02 & 51.21 & 51.21 & 47.47 & 47.45 & 63.745 & 59.590 \\
     & MART & \underline{77.93} & 53.00 & 46.70 & \underline{62.315} & \underline{58.402} & 75.44 & 75.95 & 52.88 & 52.87 & 47.03 & 47.30 & 61.235 & 57.940 \\
     & Consistency-AT & 83.42 & 51.96 & 47.72 & 65.570 & 60.711 & \textbf{84.12} & 84.04 & 52.33 & 52.14 & \textbf{48.98} & 49.06 & \textbf{66.550} & \textbf{61.911} \\
     \midrule
    \multirow{4}{*}{CIFAR-100} & PGD-AT & 56.56 & 28.80 & \underline{25.02} & 40.790 & 34.693 & 58.22 & 58.58 & 29.16 & 28.57 & 24.88 & 24.23 & 41.550 & 34.862 \\
     & TRADES & 55.39 & 29.36 & 24.51 & 39.950 & 33.983 & 57.50 & 57.25 & 29.42 & 29.00 & 25.05 & 24.45 & 41.275 & 34.897 \\
     & MART & 49.83 & 30.38 & 25.00 & 37.415 & 33.295 & 50.72 & 50.72 & 30.33 & 30.33 & 25.34 & 25.06 & 38.030 & 33.796 \\
     & Consistency-AT & 58.53 & 29.28 & 25.39 & 41.960 & 35.417 & \textbf{60.33} & 60.51 & 29.97 & 29.64 & \textbf{26.31} & 25.97 & \textbf{43.320} & \textbf{36.641} \\
     \midrule
    \multirow{4}{*}{Tiny-ImageNet} & PGD-AT & 46.32 & 21.58 & 17.07 & 31.695 & 24.947 & 47.68 & 48.32 & 22.13 & 22.00 & 17.77 & 17.91 & 32.725 & 25.891 \\
     & TRADES & 46.75 & 21.52 & 16.60 & 31.675 & 24.500 & 48.77 & 48.85 & 21.98 & 21.93 & 16.92 & 16.71 & 32.845 & 25.124 \\
     & MART & 39.70 & 22.90 & 17.18 & 28.440 & 23.982 & 41.76 & 41.66 & 23.40 & 23.29 & 17.79 & 17.78 & 29.775 & 24.951 \\
     & Consistency-AT & 46.54 & 21.89 & 17.60 & 32.070 & 25.541 & \textbf{49.74} & 49.85 & 23.16 & 22.84 & \textbf{18.84} & 18.74 & \textbf{34.290} & \textbf{27.329} \\
     \bottomrule
    \end{tabular}
  }
\end{table*}

\begin{table*}[htb]
  \caption{Improvements that the proposed RPAT achieved upon four benchmarks. The results are acquired with ResNet-18 and $\ell_{2}$ norm.}
  \label{tab:l_2}
  \centering
  \renewcommand\arraystretch{0.95}
  \resizebox{17.47cm}{!}{
    \begin{tabular}{llccccccccccccc}
    \toprule
    \multicolumn{1}{c}{\multirow{3.6}{*}{Dataset}} & \multicolumn{1}{c}{\multirow{3.6}{*}{Method}} & \multicolumn{5}{c}{\textbf{Benchmarks}} & \multicolumn{8}{c}{\textbf{+RPAT (Ours)}} \\ 
    \cmidrule(r){3-7} \cmidrule(r){8-15}
    \multicolumn{1}{c}{} & \multicolumn{1}{c}{} & Clean & PGD-20 & AA & \multirow{2.3}{*}{Mean} & \multirow{2.3}{*}{NRR} & \multicolumn{2}{c}{Clean} & \multicolumn{2}{c}{PGD-20} & \multicolumn{2}{c}{AA} & \multirow{2.3}{*}{Mean} & \multirow{2.3}{*}{NRR} \\ 
    \cmidrule{3-3} \cmidrule{4-4} \cmidrule{5-5} \cmidrule{8-9} \cmidrule{10-11} \cmidrule{12-13}
    \multicolumn{1}{c}{} & \multicolumn{1}{c}{} & best & best & best &  &  & best & final & best & final & best & final &  &  \\ 
    \midrule
    \multirow{4}{*}{CIFAR-10} & PGD-AT & 87.76 & 67.92 & 66.36 & 77.060 & 75.574 & 88.20 & 88.45 & 68.56 & 68.52 & 67.63 & 67.62 & 77.915 & 76.557 \\
     & TRADES & 83.99 & 68.60 & 65.93 & 74.960 & 73.872 & 85.17 & 84.80 & 68.95 & 68.90 & 67.67 & 67.37 & 76.420 & 75.418 \\
     & MART & 84.09 & 68.32 & 66.28 & 75.185 & 74.130 & 84.65 & 84.65 & 68.60 & 68.60 & 66.69 & 66.70 & 75.670 & 74.604 \\
     & Consistency-AT & 88.76 & 69.35 & 67.46 & 78.110 & 76.658 & \textbf{89.38} & 89.52 & 70.41 & 70.36 & \textbf{69.44} & 69.55 & \textbf{79.410} & \textbf{78.158} \\
     \midrule
    \multirow{4}{*}{CIFAR-100} & PGD-AT & 65.00 & 41.54 & \underline{39.27} & 52.135 & 48.960 & 65.14 & 65.44 & 41.61 & 41.28 & 39.23 & 39.20 & 52.185 & 48.969 \\
     & TRADES & 61.25 & 43.05 & 40.15 & 50.70 & 48.505 & 62.41 & 62.64 & 43.56 & 43.10 & \textbf{40.36} & 40.12 & 51.385 & 49.020 \\
     & MART & 60.08 & 43.42 & 39.85 & 49.965 & 47.92 & 60.57 & 60.57 & 43.92 & 43.90 & \textbf{40.36} & 40.36 & 50.465 & 48.442 \\
     & Consistency-AT & 65.14 & 42.28 & 39.92 & 52.530 & 49.503 & \textbf{65.54} & 66.81 & 42.81 & 42.51 & 40.26 & 40.61 & \textbf{52.900} & \textbf{49.880} \\
     \midrule
    \multirow{4}{*}{Tiny-ImageNet} & PGD-AT & 59.36 & 43.12 & 40.90 & 50.130 & 48.431 & 60.28 & 60.28 & 44.06 & 44.06 & 41.89 & 41.87 & 51.085 & 49.430 \\
     & TRADES & 57.61 & 44.70 & 42.33 & 49.970 & 48.802 & 58.46 & 58.50 & 44.87 & 44.65 & 42.42 & 42.44 & 50.440 & 49.165 \\
     & MART & 56.47 & 44.84 & 42.22 & 49.345 & 48.316 & 57.17 & 56.95 & 45.02 & 44.67 & 42.51 & 42.16 & 49.840 & 48.762 \\
     & Consistency-AT & 61.49 & 44.92 & 42.59 & 52.040 & 50.324 & \textbf{62.19} & 62.64 & 45.28 & 44.87 & \textbf{43.17} & 42.85 & \textbf{52.680} & \textbf{50.963} \\
    \bottomrule
    \end{tabular}
  }
\end{table*}

\subsection{Evaluation Measures}\label{subsec:exp_measure}

There are five measures in total adopted to evaluate the experimental methods from three different perspectives. Firstly, we report clean accuracy on benign test data, which reflects the ability of the robust models to maintain performance in benign cases. Secondly, we evaluate adversarial robustness under two adversarial attack methods, namely PGD-20~\cite{madry2018towards} and Auto-Attack (AA)~\cite{croce2020reliable}, with default settings aligning with Appendix~\ref{subsec:exp_setup} and random starts. Finally, to test the performance \textit{w.r.t.} the current accuracy-robustness trade-off problem in AT, on the one hand, we report the mean of the clean accuracy and the robust score under Auto-Attack acquired above, and on the other hand, we introduce the Natural-Robustness Ratio (NRR)~\cite{gowda2024conserve}:
\begin{equation*}
\textit{NRR} = \frac{2 \times \textit{Clean Accuracy} \times \textit{Adversarial Robustness}}{\textit{Clean Accuracy} + \textit{Adversarial Robustness}},
\end{equation*}
which emphasizes how well the robust model strikes a balance between clean accuracy and robustness. A lower NRR value suggests that the model might prioritize one side over the other. All the results reported are averages of three runs, with the ``best'' performance of each run recorded on the best checkpoint achieving the highest PGD-20 accuracy, which ensures that the training process is isolated from AA, the test-time adversary to evaluate the robustness.

\begin{figure*}[ht]
\begin{minipage}{0.6\linewidth}
\captionof{table}{Comparison of the proposed RPAT$^{++}$ with current SOTAs on accuracy-robustness trade-off with PreActResNet-18. The best results are marked in bold.}
\label{tab:SOTAs_PRN18}
\centering
\renewcommand\arraystretch{1.1}
\resizebox{10.8cm}{!}{
    \begin{tabular}{cllcccccccc}
    \toprule
    \multirow{2.3}{*}{Norm} & \multicolumn{2}{l}{\multirow{2.3}{*}{Method}} & \multicolumn{4}{c}{CIFAR-10} & \multicolumn{4}{c}{CIFAR-100} \\
    \cmidrule(r){4-7} \cmidrule(r){8-11}
     & \multicolumn{2}{l}{} & Clean & AA & Mean & NRR & Clean & AA & Mean & NRR \\
     \midrule
     \midrule
    \multirow{13}{*}{$\ell_{\infty}$} & WA & {\scriptsize UAI'18} & 83.50 & 49.89 & 66.695 & 62.461 & 57.26 & 25.83 & 41.545 & 35.601 \\
     & MMA & {\scriptsize ICLR'20} & \textbf{85.50} & 37.20 & 61.350 & 51.844 & \textbf{60.60} & 18.40 & 39.500 & 28.229 \\
     & AWP & {\scriptsize NeurIPS'20} & 81.11 & 50.09 & 65.600 & 61.933 & 54.10 & 25.16 & 39.630 & 34.347 \\
     & GAIRAT & {\scriptsize ICLR'21} & 78.70 & 37.70 & 58.200 & 50.979 & 52.00 & 19.80 & 35.900 & 28.680 \\
     & KD+SWA & {\scriptsize ICLR'21} & 84.06 & 49.82 & \textbf{66.940} & 62.562 & 57.17 & 25.66 & 41.415 & 35.422 \\
     & EWAT & {\scriptsize ICML'21} & 82.80 & 48.20 & 65.500 & 60.931 & 54.20 & 23.52 & 38.860 & 32.805 \\
     & MAIL & {\scriptsize NeurIPS'21} & 79.50 & 39.60 & 59.550 & 52.867 & 46.50 & 16.70 & 31.600 & 24.574 \\
     & TE & {\scriptsize ICLR'22} & 82.04 & 50.12 & 66.080 & 62.225 & 56.41 & 25.84 & 41.125 & 35.444 \\
     & SOVR & {\scriptsize ICML'23} & 81.90 & 49.40 & 65.650 & 61.628 & 52.10 & 24.30 & 38.200 & 33.142 \\
     & ReBAT & {\scriptsize NeurIPS'23} & 82.09 & 50.72 & 66.405 & 62.700 & 56.13 & 27.60 & 41.865 & 37.004 \\
     & \cellcolor{black!10}\textbf{RPAT$^{++}$} & \cellcolor{black!10}{\scriptsize \textbf{Ours}} & \cellcolor{black!10}82.63 & \cellcolor{black!10}\textbf{51.00} & \cellcolor{black!10}66.815 & \cellcolor{black!10}\textbf{63.072} & \cellcolor{black!10}56.84 & \cellcolor{black!10}\textbf{27.68} & \cellcolor{black!10}\textbf{42.260} & \cellcolor{black!10}\textbf{37.230} \\
     \midrule
     \midrule
    \multirow{2}{*}{$\ell_{2}$} & ReBAT & {\scriptsize NeurIPS'23} & 88.79 & 71.00 & 79.895 & 78.905 & 65.58 & 42.67 & 54.125 & 51.701 \\
    & \cellcolor{black!10}\textbf{RPAT$^{++}$} & \cellcolor{black!10}{\scriptsize \textbf{Ours}} & \cellcolor{black!10}\textbf{89.06} & \cellcolor{black!10}\textbf{71.26} & \cellcolor{black!10}\textbf{80.160} & \cellcolor{black!10}\textbf{79.172} & \cellcolor{black!10}\textbf{65.63} & \cellcolor{black!10}\textbf{42.85} & \cellcolor{black!10}\textbf{54.240} & \cellcolor{black!10}\textbf{51.848} \\
     \bottomrule
    \end{tabular}
  }
\end{minipage} \hspace{1.2em}
\begin{minipage}{0.36\linewidth}
\captionof{table}{Comparison of RPAT$^{++}$ with SOTAs on CIFAR-10 with WideResNet-34-10 and $\ell_{\infty}$ norm.}
\label{tab:SOTAs_WRN}
\centering
\renewcommand\arraystretch{1.095}
\resizebox{6.24cm}{!}{
    \begin{tabular}{llcccc}
    \toprule
    \multicolumn{2}{l}{Method} & Clean & AA & Mean & NRR \\
    \midrule
    WA & {\scriptsize UAI'18} & 87.66 & 52.65 & 70.155 & 65.787 \\
    MMA & {\scriptsize ICLR'20} & \textbf{87.80} & 43.10 & 65.450 & 57.818 \\
    AWP & {\scriptsize NeurIPS'20} & 85.63 & 53.32 & 69.475 & 65.718 \\
    GAIRAT & {\scriptsize ICLR'21} & 83.00 & 41.80 & 62.400 & 55.599 \\
    KD+SWA & {\scriptsize ICLR'21} & 87.45 & 53.59 & 70.520 & 66.456 \\
    EWAT & {\scriptsize ICML'21} & 86.00 & 51.60 & 68.800 & 64.500 \\
    MAIL & {\scriptsize NeurIPS'21} & 82.20 & 43.30 & 62.750 & 56.721 \\
    TE & {\scriptsize ICLR'22} & 85.97 & 52.88 & 69.425 & 65.482 \\
    SOVR & {\scriptsize ICML'23} & 85.00 & 53.10 & 69.050 & 65.366 \\
    ReBAT & {\scriptsize NeurIPS'23} & 85.25 & 54.78 & 70.015 & 66.700 \\
    ADR & {\scriptsize ICLR'24} & 84.67 & 53.25 & 68.960 & 65.381 \\
    CURE & {\scriptsize ICLR'24} & 87.05 & 52.10 & 69.575 & 65.186 \\
    \rowcolor{black!10} \textbf{RPAT$^{++}$} & {\scriptsize \textbf{Ours}} & 86.76 & \textbf{54.97} & \textbf{70.865} & \textbf{67.300} \\
    \midrule
    ReBAT* & {\scriptsize NeurIPS'23} & 86.66 & 55.64 & 71.150 & 67.769 \\
    \rowcolor{black!10} \textbf{RPAT$^{++}$*} & {\scriptsize \textbf{Ours}} & \textbf{87.57} & \textbf{55.79} & \textbf{71.680} & \textbf{68.158} \\
    \bottomrule
    \end{tabular}
  }
\end{minipage}
\end{figure*}

\begin{figure}[htbp]
  \centering
  \subfloat[\small Impact of loss function.]
  {\includegraphics[width=0.222\textwidth]{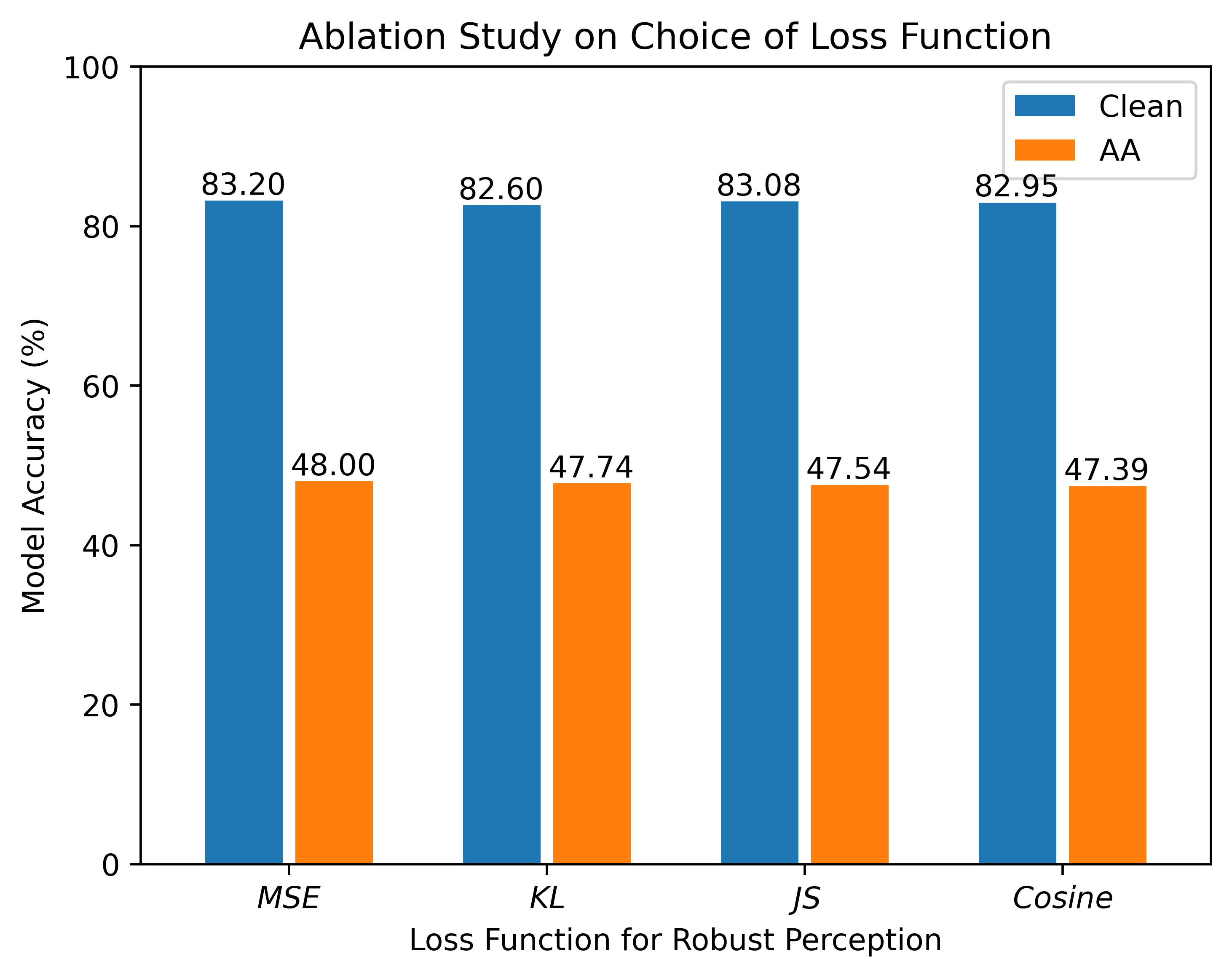}}
  \subfloat[\small Impact of hyper-parameter $\alpha$.]
  {\includegraphics[width=0.25\textwidth]{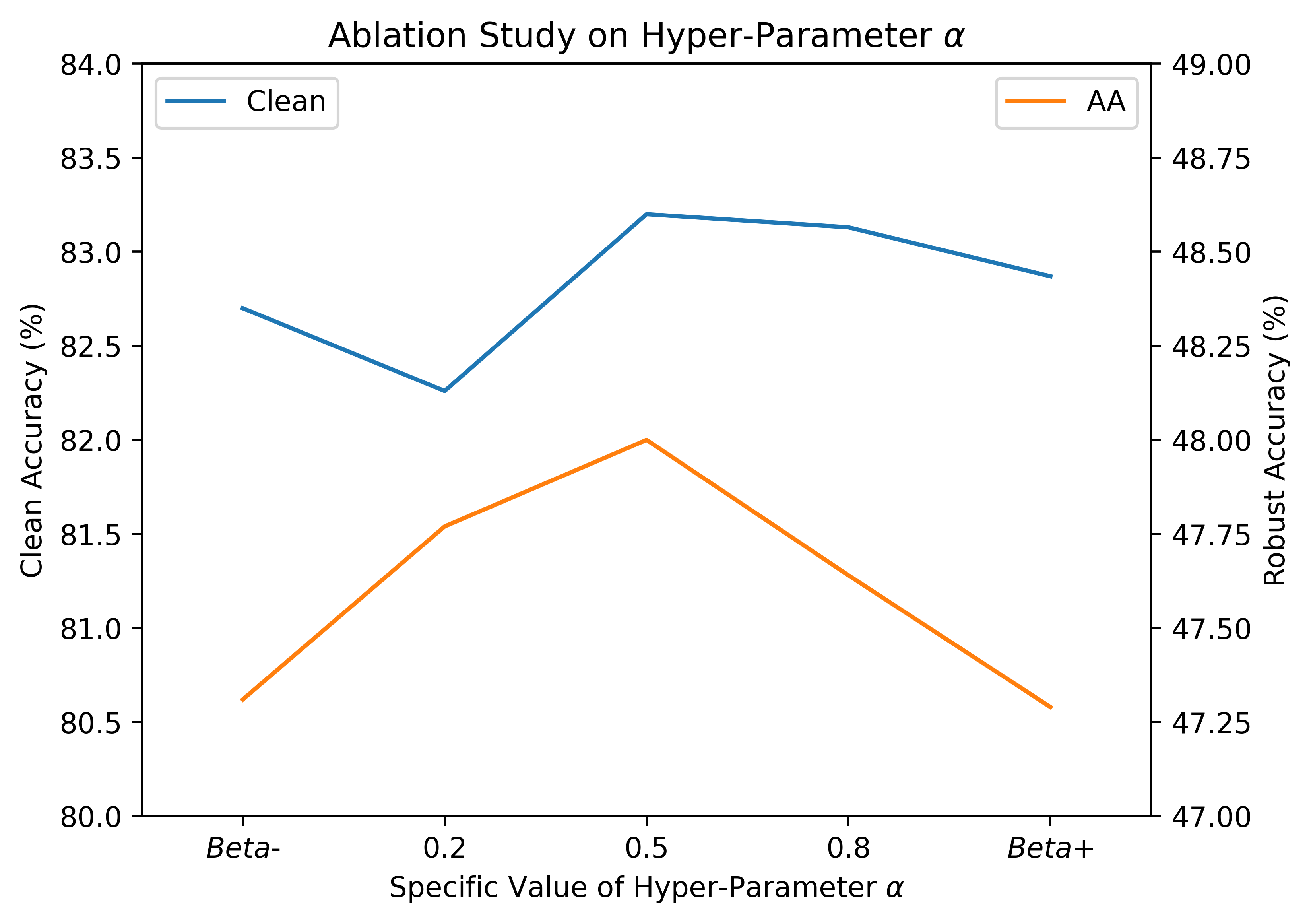}}
  \caption{Two ablation studies of the proposed RPAT on CIFAR-10 with ResNet-18 under $\ell_{\infty}$ norm. The \textit{MSE} loss in (a) and the points with $\alpha = 0.5$ in (b) correspond to the record of ``PGD-AT+RPAT'' on CIFAR-10 in Table~\ref{tab:l_inf}.}
  \label{fig:ablation}
\end{figure}

\subsection{Effectiveness in Improving Baselines}\label{subsec:exp_baselines}

The proposed RPAT can be easily implemented with existing AT methods by simply appending the \textit{Robust Perception} term. In this section, we illustrate the improvements that RPAT can bring to our four experimental benchmarks, namely PGD-AT, TRADES, MART, and Consistency-AT, on CIFAR-10, CIFAR-100, and Tiny-ImageNet with ResNet-18, demonstrating the wide effectiveness of our method. The results under $\ell_{\infty}$ and $\ell_{2}$ threat models are shown in Table~\ref{tab:l_inf} and Table~\ref{tab:l_2}, respectively, where the best clean accuracy, AA robustness, as well mean and NRR for their trade-off on each of the datasets are marked in bold, while individual scores not improved are marked with underline. Note that PGD-20 is the indicator for selecting the ``best'' checkpoint as explained in Section~\ref{subsec:exp_measure}, and the ``final'' scores are additionally provided to show that RPAT does not suffer from serious robust-overfitting, which is also evidence for appropriate algorithm design. Thus, these fields are not involved in the comparison. It can be found that RPAT achieves significant improvement for most scores among the tables, where Consistency-AT+RPAT is basically the most competitive for all the experimental groups, which demonstrates the general effectiveness of our method in mitigating the current trade-off problem.

\subsection{Advancement beyond 12 SOTAs}\label{subsec:exp_SOTAs}

To demonstrate the advancement that the proposed RPAT contributes, we further integrate RPAT into ReBAT~\cite{wang2024balance}, one of the current SOTA AT methods on the accuracy-robustness trade-off problem. It implements WA~\cite{izmailov2018averaging}, averaging multiple points along the SGD trajectory for better generalization, through the exponential moving average (EMA)~\cite{rebuffi2021data} strategy with a decay rate of 0.999. Thus, we also involve WA itself (\ie, with PGD-AT), as well as KD+SWA~\cite{chen2021robust} which shares a similar idea, for comparison. Other differences in ReBAT include \textit{KL loss} and a reduced learning rate decay factor from 0.1 to 0.5. We refer to our method upon ReBAT as RPAT$^{++}$, and compare it with 12 previous SOTAs introduced in Section~\ref{subsec:backg_tradeoff}, respectively with PreActResNet-18 and WideResNet-34-10 as shown in Table~\ref{tab:SOTAs_PRN18} and Table~\ref{tab:SOTAs_WRN}. The clean and AA scores of the SOTAs are from the best checkpoint among their original papers, Kanai et al.~\cite{kanai2023one}, Wang et al.~\cite{wang2024balance}, and our reproduction if any (so due to the lack of previous record, ADR and CURE are excluded for PreActResNet-18). Beyond the basic CIFAR-10 results under $\ell_{\infty}$ norm, Table~\ref{tab:SOTAs_PRN18} involves CIFAR-100 and $\ell_{2}$ norm, and Table~\ref{tab:SOTAs_WRN} further consider the \textit{Cutmix}~\cite{yun2019cutmix} data augmentation strategy as marked by ``*'', which ensures the comprehensiveness of our comparison. The results show that the proposed RPAT$^{++}$ achieves new SOTA performance for most of the measures beyond previous SOTAs on the accuracy-robustness trade-off problem, which strongly supports the contribution of this work.

\subsection{Ablation Study}\label{subsec:exp_ablation}

As the proposed RPAT can be implemented upon existing AT methods by simply appending the \textit{Robust Perception} term into their original objectives, the comparison in performance between and after adding RPAT in Section~\ref{subsec:exp_baselines} and Section~\ref{subsec:exp_SOTAs} have achieved a comprehensive ablation study on the effectiveness of RPAT. Therefore, in this section, we mainly investigate the impact of two setting details, respectively the loss function adopted for the \textit{Robust Perception} term (Figure~\ref{fig:ablation} (a)) and the specific value for the hyper-parameter $\alpha$ (Figure~\ref{fig:ablation} (b)). Besides, we further discuss an optional trick suggested by ReBAT but excluded in the above experiments in Appendix~\ref{subsec:exp_SOTAs}.

Specifically, just as we expected, different loss functions to calculate the similarity between model perceptions, including \textit{MSE}, \textit{KL}, \textit{JS} and \textit{Cosine}, would not significantly impact the final performance of RPAT, which suggests that the effectiveness of our method is from the successful conceptual design of \textit{Robust Perception}, instead of relying on specific implementation of the regularization term. Then, for $\alpha$, we try 0.2 and 0.8 besides the default value 0.5 on the one hand, and also introduce \textit{beta distribution} $\beta \sim Beta(0.75, 0.75)$ as suggested by Zhang et al.~\cite{zhang2019mixup} on the other hand. Specifically, we try $\alpha = \min(\beta, (1 - \beta))$ and $\alpha = \max(\beta, (1 - \beta))$, namely indicated as ``\textit{Beta}-'' and ``\textit{Beta}+'' in (Figure~\ref{fig:ablation} (b)), as they respectively approach adversarial and benign samples more. The results show that the default $\alpha = 0.5$ is an appropriate choice, which is reasonable because the midpoint is expected to be more representative for the ideal target that \textit{Robust Perception} holds for all $\alpha \in [0, 1]$.

\section{Conclusion}\label{sec:conclu}

This work studies the current accuracy-robustness trade-off problem in AT. To the best of our knowledge, this is the first work that explicitly attributes the trade-off problem to the over-sufficient (rather than insufficient) learning of hard adversarial samples. We surprisingly reveal that such hard samples are already learned better than those successfully defended in model perception, which means the conventional AT objective is no longer effective to further optimize the model \textit{w.r.t.} these hard samples. Instead, it forces the model to ignore the perturbations, in turn degrading their effectiveness in supporting the establishment of decision boundary to an appropriate location. To mitigate this, our new AT objective, \textit{Robust Perception}, encourages smoother perception transition towards the decision boundary along with perturbations, based on which the proposed RPAT method achieves new SOTA performance. Still, exploring better methods to approximate this ideal objective would be a promising future direction.

\section*{Acknowledgements}

This work was supported by the National Natural Science Foundation of China (No. 62471420), GuangDong Basic and Applied Basic Research Foundation (2025A15150122 96), and CCF-Tencent Rhino-Bird Open Research Fund.

{
    \small
    \bibliographystyle{ieeenat_fullname}
    \bibliography{main}
}
\clearpage
\onecolumn

\appendix

\section*{\centering{\Large Appendix}}

{\centering{\large Failure Cases Are Better Learned But Boundary Says Sorry: Facilitating Smooth Perception Change for Accuracy-Robustness Trade-Off in Adversarial Training \\ \vspace{0.5em}}}

\section{Experimental Setup}\label{subsec:exp_setup}

We employ common experimental settings aligned with previous AT works. For the outer minimization, we adopt SGD optimizer with momentum 0.9, batch size 128, weight decay $5 \!\times\! 10^{-4}$ and initial learning rate 0.1. In Section~\ref{subsec:exp_baselines}, following Pang et al.~\cite{pang2021bag}, which explores various settings for AT, we train 110 epochs with learning rate decay by a factor of 0.1 at 100 and 105 epochs for better experimental efficiency. While in Section~\ref{subsec:exp_SOTAs}, we adopt 200 training epochs with learning rate decay at 100 and 150 epochs to ensure fairness in comparison with the current SOTAs that also use this default setting. For the inner maximization, under the $\ell_{\infty}$ threat model with perturbation budget $\epsilon = 8/255$, we employ PGD-10 adversary with step size $\alpha = 2/255$ and maximum optimization step 10, except for TRADES which crafts adversarial samples through maximizing its \textit{KL} regularization term~\cite{zhang2019theoretically}. While under the $\ell_2$ threat model with maximal perturbation budget $\epsilon = 128/255$, we have step size $\alpha = 32/255$. In the main experiments, following their original papers, we set the regularization parameter $\lambda = 6$ for TRADES and MART while $\lambda = 1$ for Consistency-AT and ours, and fix our specific hyper-parameter $\alpha=0.5$. Other values of these parameters are further considered by our ablation studies in Section~\ref{subsec:exp_ablation}. Then, for the data pre-processing, we normalize benign images into [0, 1], and employ standard data augmentations, including random crop with 4-pixel zero padding and random horizontal flip with 50\% of probability. 

\vspace{1ex} \noindent The experiments are conducted on Ubuntu 22.04 OS with Intel Xeon Gold 6226R $\!$@$\!$ 2.90GHz CPU, 512GB RAM and 8 $\!\times\!$ NVIDIA GeForce RTX 3090 GPUs, and are implemented with Python 3.8.19 and PyTorch 1.11.0 $\!$+$\!$ cu113.

\section{Proof of Theorems}

In this section, we supplement the proofs of Theorem 1 and Theorem 2 in Section~\ref{subsec:objective}, which respectively demonstrate the effectiveness of the proposed RPAT from the perspectives of local linearity~\cite{goodfellow2015explaining} and \textit{Lipschitz} regularization~\cite{krishnan2020lipschitz,pauli2021training,fazlyab2023certified,abuduweili2024estimating}.

\subsection{Proof of Theorem 1}\label{app:theorem_1}

\noindent \textbf{Theorem 1} (Section~\ref{subsec:objective}) \textbf{.} \textit{Let H be the Hessian Matrix such that $H_{h_{\bm{\theta}}}(\mathbf{x})$ $= \nabla_\mathbf{x}^2h_{\bm{\theta}}(\mathbf{x})$, then with the new optimization objective of Robust Perception, we have:}
\begin{equation}
\forall \, \Delta, \,\, \Delta^\top \!\cdot H_{h_{\bm{\theta}}}(\mathbf{x}) \cdot \Delta \to 0.
\end{equation}

\noindent \textit{Proof.} 

\vspace{1ex} \noindent Based on Definition 1 in Section~\ref{subsec:objective}, for any robust model $\bm{\theta}$ satisfies the proposed \textit{Robust Perception}, we have:
\begin{equation}\label{eq:5}
\forall \, \alpha \in [0, 1], \,\,\,  h_{\bm{\theta}}(\mathbf{x} + \alpha \cdot \Delta) - h_{\bm{\theta}}(\mathbf{x}) = \alpha \cdot (h_{\bm{\theta}}(\mathbf{x} + \Delta) - h_{\bm{\theta}}(\mathbf{x})) .
\end{equation}

\noindent Expand the $h_{\bm{\theta}}(\mathbf{x} + \alpha \cdot \Delta)$ term of Equation~(\ref{eq:5}) as a \textit{Taylor series}:
\begin{equation}\label{eq:6}
h_{\bm{\theta}}(\mathbf{x} + \alpha \cdot \Delta) = h_{\bm{\theta}}(\mathbf{x}) + \alpha \cdot J_{h_{\bm{\theta}}}(\mathbf{x}) \cdot \Delta + \frac{\alpha^2}{2} \cdot \Delta^\top \!\cdot H_{h_{\bm{\theta}}}(\mathbf{x}) \cdot \Delta + \mathcal{O}(\alpha^3),
\end{equation}
and also expand the $h_{\bm{\theta}}(\mathbf{x} + \Delta)$ term of Equation~(\ref{eq:5}) as a \textit{Taylor series}:
\begin{equation}\label{eq:7}
h_{\bm{\theta}}(\mathbf{x} + \Delta) = h_{\bm{\theta}}(\mathbf{x}) + J_{h_{\bm{\theta}}}(\mathbf{x}) \cdot \Delta + \frac{1}{2} \cdot \Delta^\top \!\cdot H_{h_{\bm{\theta}}}(\mathbf{x}) \cdot \Delta + \mathcal{O}(\|\Delta\|^3).
\end{equation}

\noindent Substitute Equation~(\ref{eq:6}) and Equation~(\ref{eq:7}) back into Equation~(\ref{eq:5}), we have:
\begin{equation}\label{eq:8}
\forall \, \alpha \in [0, 1], \,\,\,  \alpha \cdot J_{h_{\bm{\theta}}}(\mathbf{x}) \cdot \Delta + \frac{\alpha^2}{2} \cdot \Delta^\top \!\cdot H_{h_{\bm{\theta}}}(\mathbf{x}) \cdot \Delta + \mathcal{O}(\alpha^3) = \alpha \cdot (J_{h_{\bm{\theta}}}(\mathbf{x}) \cdot \Delta + \frac{1}{2} \cdot \Delta^\top \!\cdot H_{h_{\bm{\theta}}}(\mathbf{x}) \cdot \Delta + \mathcal{O}(\|\Delta\|^3)),
\end{equation}
which can be then simplified as:
\begin{equation}\label{eq:9}
\forall \, \alpha \in [0, 1], \,\,\,  (\frac{\alpha^2}{2} - \frac{\alpha}{2}) \cdot \Delta^\top \!\cdot H_{h_{\bm{\theta}}}(\mathbf{x}) \cdot \Delta + \mathcal{O}(\alpha^3) - \alpha \cdot \mathcal{O}(\|\Delta\|^3)) = 0.
\end{equation}

\noindent Equation~(\ref{eq:9}) directly means:
\begin{equation}\label{eq:10}
\Delta^\top \!\cdot H_{h_{\bm{\theta}}}(\mathbf{x}) \cdot \Delta = 0, \,\,\, \mathcal{O}(\alpha^3) = 0, \, \text{ and } \,\, \mathcal{O}(\|\Delta\|^3)) = 0,
\end{equation}
otherwise we can easily find a specific value of $\alpha$ except 0 and 1 that violates Equation~(\ref{eq:9}).

\vspace{0.5em} \noindent Therefore, we have Theorem 1 proven, meaning that \textit{Robust Perception} limits the second-order and higher-order nonlinear effects within the adversarial perturbation to the model perception, just as we suggested in Section~\ref{subsec:objective}.

\subsection{Proof of Theorem 2}\label{app:theorem_2}

\noindent \textbf{Theorem 2} (Section~\ref{subsec:objective}) \textbf{.} \textit{Let J be the Jacobian Matrix such that $J_{h_{\bm{\theta}}}(\mathbf{x})$ $= \nabla_\mathbf{x} h_{\bm{\theta}}(\mathbf{x})$, then with the new optimization objective of Robust Perception, we have:}
\begin{equation}\label{eq:11}
\forall \, \alpha \in [0,1], \,\, J_{h_{\bm{\theta}}}(\mathbf{x} + \alpha \cdot \Delta) \to J_{h_{\bm{\theta}}}(\mathbf{x}),
\end{equation}
\textit{then with $\| \cdot \|_{\text{spec}}$ denoting the Spectral Norm and $\gamma$ being any micro value, given $\| J_{h_{\bm{\theta}}}(\mathbf{x} + \alpha \cdot \Delta) - J_{h_{\bm{\theta}}}(\mathbf{x}) \|_{\mathrm{spec}} \leq \gamma$, the function $h_{\bm{\theta}}(\mathbf{x})$ can be referred to as $K$-Lipschitz with the Lipschitz constant $K$ upper-bounded by:}
\begin{equation}\label{eq:12}
K \,\leq\, \sup_\mathbf{x} \| J_{h_{\bm{\theta}}}(\mathbf{x}) \|_{\mathrm{spec}} + \gamma.
\end{equation}

\noindent \textit{Proof.} 

\vspace{1ex} \noindent Provided the result in Equation~(\ref{eq:10}), we can simplify the \textit{Taylor series} at $\mathbf{x}$ in Equation~(\ref{eq:6}) as:
\begin{equation}\label{eq:13}
h_{\bm{\theta}}(\mathbf{x} + \alpha \cdot \Delta) \approx h_{\bm{\theta}}(\mathbf{x}) + \alpha \cdot J_{h_{\bm{\theta}}}(\mathbf{x}) \cdot \Delta.
\end{equation}

\noindent For $h_{\bm{\theta}}(\mathbf{x} + \Delta)$, this time we expand it at $\mathbf{x} + \alpha \cdot \Delta$ (\ie, this is the point where the derivatives are considered), such that the variable here becomes $\Delta - \alpha \cdot \Delta = (1 - \alpha) \cdot \Delta$, with which we have:
\begin{equation}\label{eq:14}
h_{\bm{\theta}}(\mathbf{x} + \Delta) \approx h_{\bm{\theta}}(\mathbf{x} + \alpha \cdot \Delta) + (1 - \alpha) J_{h_{\bm{\theta}}}(\mathbf{x} + \alpha \cdot \Delta) \cdot \Delta,
\end{equation}
in which the higher-order terms are also ignored based on Equation~(\ref{eq:10}).

\vspace{1ex} \noindent Then substitute Equation~(\ref{eq:13}) into Equation~(\ref{eq:14}), we have:
\begin{equation}\label{eq:15}
h_{\bm{\theta}}(\mathbf{x} + \Delta) \approx h_{\bm{\theta}}(\mathbf{x}) + \alpha \cdot J_{h_{\bm{\theta}}}(\mathbf{x}) \cdot \Delta + (1 - \alpha) J_{h_{\bm{\theta}}}(\mathbf{x} + \alpha \cdot \Delta) \cdot \Delta.
\end{equation}

\noindent Recall the definition of \textit{Robust Perception} in Equation~(\ref{eq:5}), substitute Equation~(\ref{eq:13}) and Equation~(\ref{eq:15}) into it, we have:
\begin{equation}\label{eq:16}
\forall \, \alpha \in [0, 1], \,\,\,  \alpha \cdot J_{h_{\bm{\theta}}}(\mathbf{x}) \cdot \Delta \approx \alpha \cdot (\alpha \cdot J_{h_{\bm{\theta}}}(\mathbf{x}) \cdot \Delta + (1 - \alpha) J_{h_{\bm{\theta}}}(\mathbf{x} + \alpha \cdot \Delta) \cdot \Delta),
\end{equation}
which can be directly simplified as:
\begin{equation}\label{eq:17}
\forall \, \alpha \in [0, 1], \,\,\,  J_{h_{\bm{\theta}}}(\mathbf{x}) \approx J_{h_{\bm{\theta}}}(\mathbf{x} + \alpha \cdot \Delta).
\end{equation}

\noindent This result tells us the proposed \textit{Robust Perception} encourages the stability in \textit{Jacobian} along with the adversarial perturbation, as given in Equation~(\ref{eq:11}) of Theorem 2.

\vspace{1ex} \noindent Based on Equation~(\ref{eq:17}), let us assume that the change in \textit{Jacobian} along with the perturbation satisfies:
\begin{equation}\label{eq:18}
\| J_{h_{\bm{\theta}}}(\mathbf{x} + \alpha \cdot \Delta) - J_{h_{\bm{\theta}}}(\mathbf{x}) \|_{\mathrm{spec}} \,\leq\, \gamma,
\end{equation}
where $\gamma$ is a micro value and \textit{Spectral Norm} $\| \cdot \|_{\text{spec}}$ indicates the maximum singular value of the matrix. 

\vspace{1ex} \noindent Then, according to the triangular inequality of \textit{Spectral Norm}, we have:
\begin{equation}\label{eq:19}
\| J_{h_{\bm{\theta}}}(\mathbf{x} + \alpha \cdot \Delta)\|_{\mathrm{spec}} - \|J_{h_{\bm{\theta}}}(\mathbf{x}) \|_{\mathrm{spec}} \,\leq\, \| J_{h_{\bm{\theta}}}(\mathbf{x} + \alpha \cdot \Delta) - J_{h_{\bm{\theta}}}(\mathbf{x}) \|_{\mathrm{spec}} \,\leq\, \gamma.
\end{equation}

\noindent Since the local \textit{Lipschitz} constant $K_{\text{local}}(\cdot)$ can be denoted with \textit{Jacobian} as $\| J_{h_{\bm{\theta}}}(\cdot) \|_{\mathrm{spec}}$~\cite{miyato2018spectral}, we further have:
\begin{equation}\label{eq:20}
K_{\text{local}}(\mathbf{x} + \alpha \cdot \Delta) - \| J_{h_{\bm{\theta}}}(\mathbf{x}) \|_{\mathrm{spec}} \,\leq\, \gamma,
\end{equation}

\clearpage

\noindent with which we can finally represent the upper bound of the global \textit{Lipschitz} constant $K$ under the adversarial perturbation as:
\begin{equation}\label{eq:21}
K = \sup_\mathbf{x} K_{\text{local}}(\mathbf{x} + \alpha \cdot \Delta) \,\leq\, \sup_\mathbf{x} \| J_{h_{\bm{\theta}}}(\mathbf{x}) \|_{\mathrm{spec}} + \gamma,
\end{equation}
just as Equation~(\ref{eq:12}) of Theorem 2.

\vspace{1ex} \noindent With Theorem 2, we can refer to the function $h_{\bm{\theta}}(\mathbf{x})$ learned under \textit{Robust Perception} as $K$-\textit{Lipschitz} with an upper-bounded global \textit{Lipschitz} constant, which is expected to limit the complexity of the decision boundary as suggested in Section~\ref{subsec:objective}.

\section{Additional Results}

This section supplements more experimental results to further support our ideas and statements in this work, including more empirical evidence for our motivation in Appendix~\ref{app:motivation}, different options of the proxy for model perception in Appendix~\ref{app:proxy}, and further comparison with the current SOTA in Appendix~\ref{app:stronger}.

\subsection{More Empirical Evidence for Motivation}\label{app:motivation}

\begin{figure}[htbp]
  \centering
  \subfloat[\small Results on CIFAR-100 with PreActResNet-18.]
  {\includegraphics[width=0.4\textwidth]{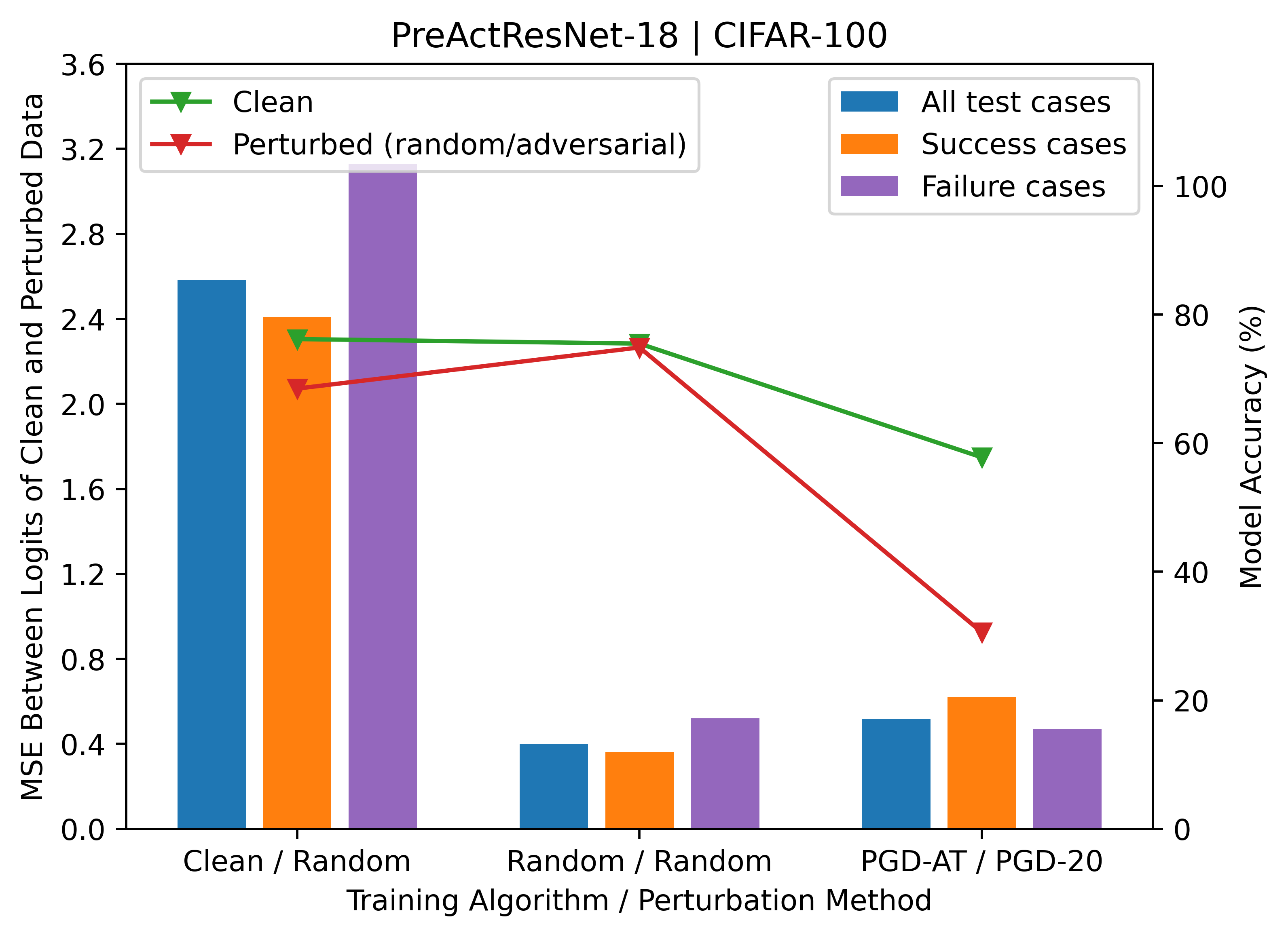}}
  \hspace{1.5em}
  \subfloat[\small Results on Tiny-ImageNet with WideResNet-34-10.]
  {\includegraphics[width=0.4\textwidth]{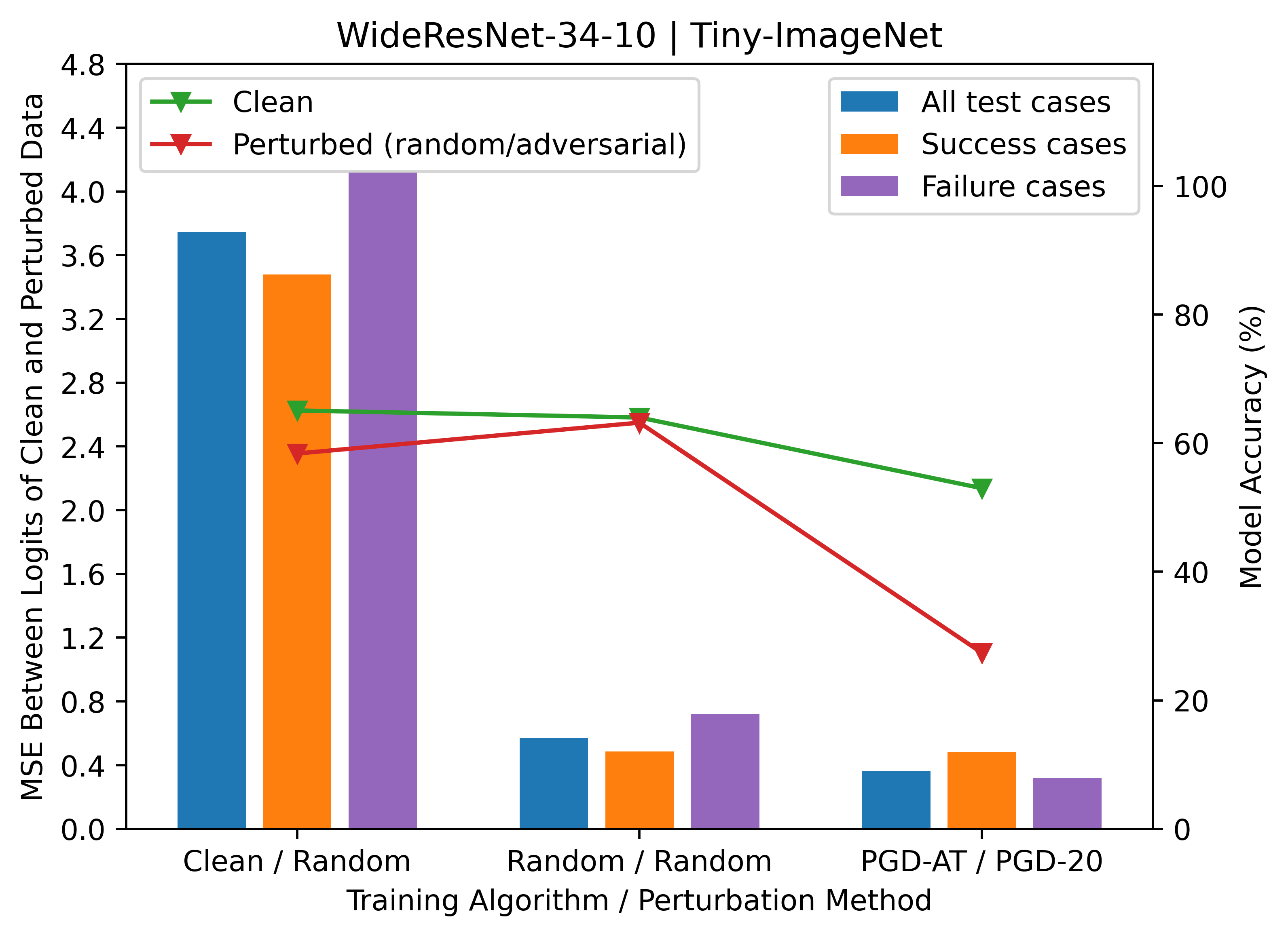}}
  \caption{More empirical evidence for our motivation, aligned with the proof-of-concept results in Figure~\ref{fig:motivation}.}
  \label{fig:app_motivation}
\end{figure}

Corresponding to the proof-of-concept experiment illustrated in Figure~\ref{fig:motivation}, which is conducted on CIFAR-10 with ResNet-18, we provide more empirical evidence respectively on CIFAR-100 with PreActResNet-18 and Tiny-ImageNet with WideResNet-34-10, as illustrated in Figure~\ref{fig:app_motivation}, both of which show similar patterns to Figure~\ref{fig:motivation}.

\subsection{Various Proxies for Model Perception}\label{app:proxy}

Regarding the proxy of model perception in our RPAT, embeddings from other layers than \textit{logits} could also be alternatives for comparing perception consistency. In this section, we further provide the results with two additional proxies, respectively the embeddings from the second-to-last layer and the third-to-last layer, as demonstrated in Table~\ref{tab:app_proxy}. It can be found that there is no significant difference in the final performance, which also reflects the universality and stability of our new AT objective. Thus, for simplicity, we uniformly utilize \textit{logits} as our proxy in the main experiments. 

\begin{table}[h!]
    \caption{Comparison of different proxies for model perception in calculating the perception consistency. The results are acquired on CIFAR-10 with ResNet-18 and $\ell_{\infty}$ norm.}
    \label{tab:app_proxy}
    \centering
    \renewcommand\arraystretch{1.07}
    \begin{tabular}{clcccc}
    \toprule
    \multicolumn{2}{c}{\textbf{Method / Proxy}} & \textbf{Clean} & \textbf{AA} & \textbf{Mean} & \textbf{NRR} \\
    \midrule
    \multicolumn{2}{c}{PGD-AT (\ie, the baseline as in Table~\ref{tab:l_inf})} & 82.92 & 46.74 & 64.83 & 59.78 \\
    \midrule
    \multirow{2}{*}{\textbf{+ RPAT}} & \textit{logits} (\ie, the current proxy as in Table~\ref{tab:l_inf}) & \textbf{83.20} & 48.00 & \textbf{65.60} & \textbf{60.88} \\
    \multirow{2}{*}{\textbf{(Ours)}} & Embedding from the second-to-last layer & 83.19 & 47.89 & 65.54 & 60.79 \\
     & Embedding from the third-to-last layer & 83.14 & \textbf{48.02} & 65.58 & \textbf{60.88} \\
    \bottomrule
    \end{tabular}
\end{table}

\subsection{Further Comparison with Current SOTA}\label{app:stronger}

Except from the ones considered in Section~\ref{subsec:exp_SOTAs}, ReBAT also suggests another additional training strategy, which is to utilize a stronger training-time adversary with larger perturbation budgets (\eg, $\epsilon = 10/255$) after the first learning rate decay. Although this seems not completely aligned with the default fairness setting of AT, we still supplement comparison with it, as ReBAT is the current SOTA method on the accuracy-robustness trade-off problem. The additional consideration of such a stronger adversary is marked by ``\$'' in Table~\ref{tab:SOTAs_stronger}.

\begin{table}[htb]
  \caption{Comparison of the proposed RPAT$^{++}$ with the current SOTA, ReBAT, under stronger training adversary on CIFAR-10 with PreActResNet-18 and $\ell_{\infty}$ norm.}
  \label{tab:SOTAs_stronger}
  \centering
    \begin{tabular}{lcccccccc}
    \toprule
    \multirow{2.3}{*}{Method} & \multicolumn{2}{c}{Clean} & \multicolumn{2}{c}{PGD-20} & \multicolumn{2}{c}{AA} & \multirow{2.3}{*}{Mean} & \multirow{2.3}{*}{NRR} \\
    \cmidrule(r){2-7}
     & best & final & best & final & best & final &  &  \\
    \midrule
     ReBAT & 82.09 & 82.05 & 55.77 & 56.03 & 50.72 & 50.70 & 66.405 & 62.700  \\
    \rowcolor{black!10} \textbf{RPAT$^{++}$} & \textbf{82.63} & \textbf{82.76} & 56.27 & 56.02 & 51.00 & 50.71 & \textbf{66.815} & \textbf{63.072} \\
    \midrule
    ReBAT\$ & 77.57 & 78.82 & 56.79 & 56.46 & 50.91 & \textbf{51.09} & 64.240 & 61.474 \\
    \rowcolor{black!10} \textbf{RPAT$^{++}\$$} & 79.25 & 79.47 & \textbf{57.04} & \textbf{56.52} & \textbf{51.25} & \textbf{51.09} & 65.250 & 62.246 \\
    \bottomrule
    \end{tabular}
\end{table}

\vspace{1ex} \noindent The results demonstrate that, although the stronger training adversary strategy helps further improve the robustness, its destruction on the clean accuracy is more significant. As a consequence, for the Mean and NRR scores measuring the trade-off, adopting such a strategy rather leads to worse results. Therefore, we would not suggest using this strategy for the proposed RPAT method by default.

\end{document}